\documentclass[letterpaper, 10 pt, conference]{ieeeconf}  % Comment this line out if you need a4paper

\IEEEoverridecommandlockouts                              % 
\usepackage{graphicx} % for pdf, bitmapped graphics files
\usepackage{epsfig} % for postscript graphics files
\usepackage{soul,color} 
\usepackage{xcolor}
\usepackage{comment}
\usepackage{xspace}
\usepackage{mathtools}
\usepackage{booktabs}
\usepackage{multirow}
\usepackage{enumerate}
\usepackage{balance}
\usepackage{gensymb}
\usepackage{tabulary}
\usepackage{dblfloatfix}
\usepackage{layouts}
\usepackage{algorithm}
\usepackage{algpseudocode} % http://ctan.org/pkg/algorithmicx
\usepackage{cite}
\usepackage{bm}
\usepackage{hyperref}
\usepackage{svg}
\usepackage{cite}
\usepackage{caption}
\usepackage{subcaption}
\usepackage{colortbl}
\usepackage{overpic}
\usepackage{rotating}
\usepackage{afterpage}
\usepackage{float}
\usepackage{url}
\usepackage{wrapfig}
\usepackage{listings}
\usepackage{adjustbox}
\usepackage{mdwtab}
\usepackage{array}
\usepackage{setspace}
\usepackage{mathdefs}
\usepackage{sonar}
\usepackage{units}
\usepackage{makecell}
\usepackage[many]{tcolorbox}
\usepackage[none]{hyphenat}

\usepackage{diagbox}
\usepackage{placeins}

\newcommand{\spatialpoint}{\ensuremath{\mathbf{x}}}

\newcolumntype{?}{!{\vrule width 2pt}}
\definecolor{lightgreen}{rgb}{0.8,1,0.8}

\usepackage{caption}
\graphicspath{{../figs/}}

\makeatletter
\newcount\my@repeat@count
\newcommand{\myrepeat}[2]{%
  \begingroup
  \my@repeat@count=\z@
  \@whilenum\my@repeat@count<#1\do{#2\advance\my@repeat@count\@ne}%
  \endgroup
}

\title{\LARGE \bf
Neural Implicit Representation for Bathymetry Estimation using Sidescan Sonar Imagery
}

\author{{Tianxiang Lin, Sacchin Sundar, Fenghao Liu, Onur Bagoren, Katherine A. Skinner} % <-this % stops a space
\thanks{T. Lin, S. Sundar, F. Liu, O. Bagoren, K.A. Skinner are with the Robotics Department, University of Michigan, Ann Arbor, MI, USA.
        {\tt\small \{tianxian, sacchin, fenghao, obagoren, kskin\}@umich.edu}}%
\thanks{This work was supported by the Office of Naval Research grant N00014-25-1-2039. Data used for experiments was collected in prior field expeditions supported by the NOAA Ocean Exploration Program under award \#NA21OAR0110196.}
}

\begin{document}
\bstctlcite{IEEEexample:BSTcontrol}

\maketitle
\thispagestyle{empty}
\pagestyle{empty}

%%%%%%%%%%%%%%%%%%%%%%%%%%%%%%%%%%%%%%%%%%%%%%%%%%%%%%%%%%%%%%%%%%%%%%%%%%%%%%%%
\begin{abstract}

Bathymetry measurement and estimation is critical for understanding the structure of underwater environments. While multibeam echosounders (MBES) provide accurate depth measurements, their deployment on surface vessels limits spatial resolution in deep water. Sidescan sonar (SSS), in contrast, can be deployed on autonomous underwater vehicles (AUVs) to operate closer to the seafloor, enabling high-resolution acoustic imaging. However, the inherent ambiguity in measurements and the limited observation coverage of sonar datasets degrade the performance of state-of-the-art methods for underwater bathymetry estimation from SSS. In this paper, we propose a differentiable neural rendering framework for SSS that enables effective estimation of bathymetry from SSS measurements. Our approach incorporates a measurement model based on SSS geometry, explicitly accounting for its wide elevation aperture and associated ambiguities. We evaluate our method on both synthetic and real-world datasets, demonstrating improved robustness and accuracy compared to alternative approaches.

\end{abstract}

%%%%%%%%%%%%%%%%%%%%%%%%%%%%%%%%%%%%%%%%%%%%%%%%%%%%%%%%%%%%%%%%%%%%%%%%%%%%%%%%
\section{Introduction}

\noindent  
Oceans cover approximately $70.8\%$ of Earth’s surface, yet remain far less explored than terrestrial environments~\cite{Pidwirny2014fundamentals}. Improving the understanding of underwater environments requires accurate reconstruction of their geometry. Bathymetry estimation, which determines the depth of the seafloor, plays a critical role in this task, facilitating applications such as autonomous navigation~\cite{Westman2019JOE}, obstacle avoidance~\cite{Liu2026RAL}, target detection~\cite{Zhang2026OE,Sethuraman2025IJRR,Lin2023ICRA} and reconstruction~\cite{Qadri2023ICRA,Xie2024RAL,Lin2025IROS}.

Ship-based multibeam echosounders (MBES) are the standard technology for bathymetric surveys. However, the spatial resolution of depth measurements degrades with increasing distance between the sensor and seafloor, and the sensors are typically deployed on surface vessels. Ship-based surveys are expensive to operate compared to low-cost autonomous underwater vehicles (AUVs) or uncrewed platforms equipped with acoustic sensors~\mbox{\cite{Fernandes2003ICESJMS,Mayer2023IHR}}.

\begin{figure}[t!]
    \centering
    \includegraphics[width=\linewidth]{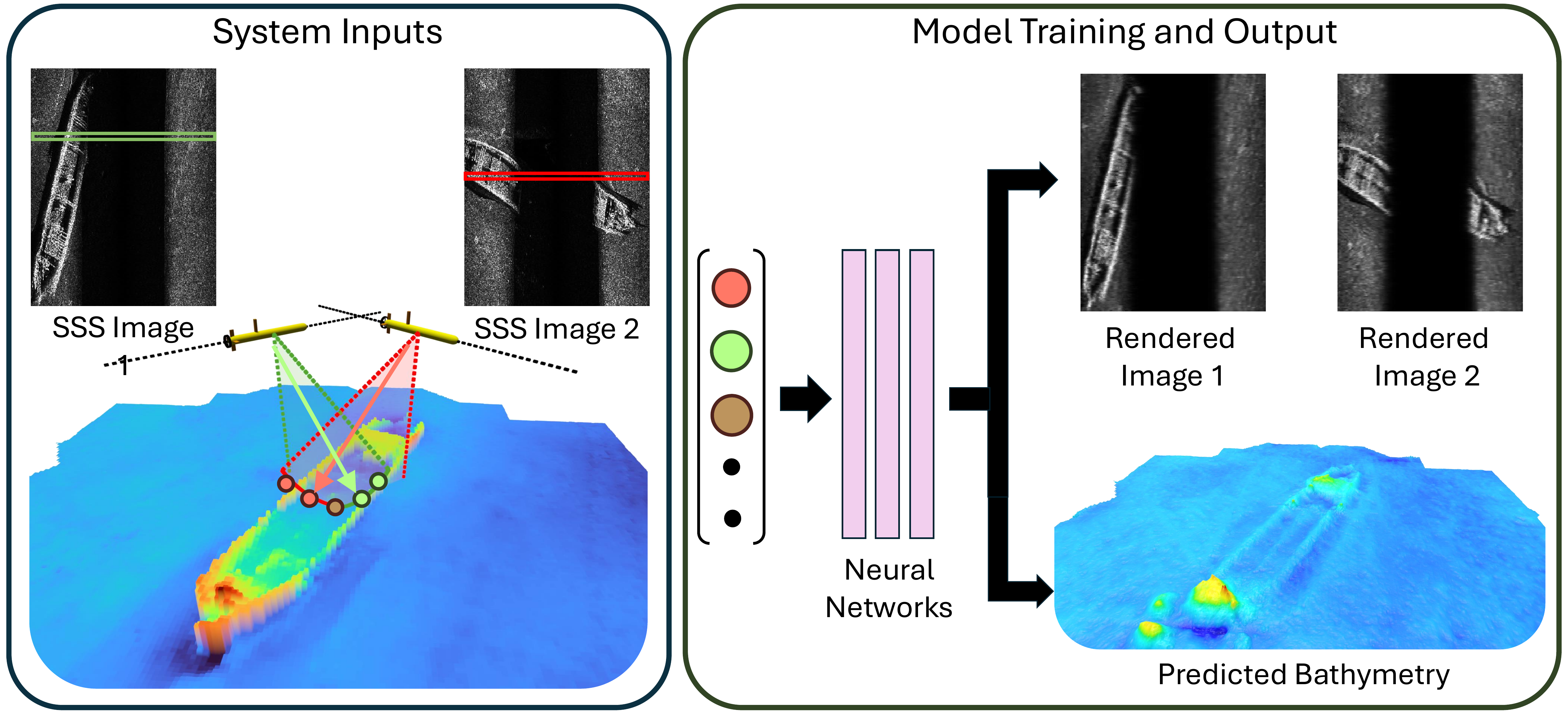}
    \caption{Overview of the sensing modality and an example bathymetry estimation task. We present example sidescan sonar (SSS) surveys of the D.M. Wilson shipwreck located in the Thunder Bay National Marine Sanctuary, together with reference bathymetry using a multibeam echosounder (MBES). Each sonar ping (highlighted by the green and red boxes) is associated with a sensor pose from the AUV. Following the SSS imaging geometry, we sample 3D points and directional vectors as inputs to the neural networks, which render SSS images and estimate the bathymetry.}
    \label{fig:overview}
    \vspace{-2em}
\end{figure}

Sidescan sonar (SSS) is compact and readily deployable on AUVs, allowing acoustic measurements to be taken closer to the seafloor to return high-resolution acoustic imagery. While some high-end SSS systems offer combined bathymetry and sidescan data~\cite{EdgeTech2205}, standard SSS does not directly provide depth measurements. The wide elevation aperture of SSS introduces an elevation ambiguity, as the object position and the number of objects along the elevation arc cannot be uniquely determined. Moreover, SSS images are view-dependent. Fully covering a survey area with multiple viewpoints is time-consuming, further increasing the difficulty of obtaining sufficient observations. At the same time, poses from state estimation are subject to noise and time-based drift. All of these factors make bathymetry estimation from SSS a challenging task. 

Neural implicit representations and differentiable rendering have recently emerged as powerful tools for 3D reconstruction from sparse observations \cite{Mildenhall2020ECCV}. By modeling geometry with neural networks, these methods enable dense reconstruction and encourage consistency across multiple views with reduced view coverage, making them suitable for providing novel view synthesis and handling the elevation ambiguity for smaller datasets. Recent work on neural implicit representations from sonar imagery has mainly focused on imaging sonars~\cite{Qadri2023ICRA, Xie2024RAL, Lin2025IROS}.
While prior studies represent bathymetry using sinusoidal representation networks (SIRENs) and render SSS measurements with a Lambertian scattering model \cite{Xie2022Sensors,Xie2025JOE}, their surface-rendering formulations assume a single seafloor intersection for each elevation arc. This assumption does not always hold over uneven terrain or near complex structures, where an elevation arc may intersect the surface multiple times.
Under insufficient view coverage, the elevation ambiguity remains unresolved, and pose noise and drift introduce conflicting constraints, causing the optimization to become ill-conditioned and prone to local minima.

To address this research gap, we propose a novel method that uses differentiable neural rendering to estimate bathymetry from SSS imagery. An overview of the proposed method is shown in Fig. \ref{fig:overview}. The specific contributions are:

\begin{itemize}
    \item We develop a novel differentiable neural rendering framework for bathymetry estimation from SSS imagery that is capable of reconstructing complex structures.
    \item We integrate a measurement model with implicit representations tailored to SSS geometry, explicitly accounting for its wide elevation aperture and inherent ambiguities.
    \item We present quantitative and qualitative validation on both synthetic and real-world datasets, demonstrating improved accuracy in bathymetry estimation and in reconstruction of complex shipwreck structures compared to existing approaches. 
\end{itemize}

%%%%%%%%%%%%%%%%%%%%%%%%%%%%%%%%%%%%%%%%%%%%%%%%%%%%%%%%%%%%%%%%%%%%%%%%%%%%%%%%
\section{Related Work}
\noindent
Bathymetric surveys largely rely on ship-based MBES, which can provide accurate and large-scale depth measurements. The Sea Beam system~\cite{Renard1979IHR} was among the first multibeam echosounders designed for real-time large-scale bathymetric mapping, utilizing a 16-beam echosounder. Recent work introduced an MBES model based on geometric optimization to better handle complex seafloor terrain~\cite{Lu2024AS}. Chen et al.~\cite{Chen2026RS} further improved the reconstruction accuracy by incorporating the denoising diffusion probabilistic model. However, the spatial resolution of MBES decreases with increasing water depth, as the sensors are typically mounted on surface vessels far from the seafloor. In addition, ship-based surveys are expensive and time-consuming to operate compared to deploying AUVs equipped with compact acoustic sensors such as SSS \cite{Fernandes2003ICESJMS,Mayer2023IHR}. As a result, MBES-based approaches are not always practical for large-scale or high-resolution underwater mapping.
In this work, we instead focus on SSS as an alternative sensing modality for bathymetry estimation.

SSS provides high-resolution acoustic imagery of underwater environments and is compatible and readily deployable on low-cost AUVs for large-scale surveys, but does not provide direct bathymetry measurement and solve its geometric ambiguity along the elevation angle. Many studies have focused on explicitly modeling the geometry and physics of acoustic image formation, including elevation estimation from 2D sonar images \cite{wang2021elevation} and spatial acoustic projection models \cite{arnold2022spatial}. Volumetric and albedo-based formulations further investigate how reflectivity and propagation effects influence 3D sonar reconstruction \cite{westman2020volumetric, westman2020theory}. While these models provide important physical insight, their reliance on carefully designed analytical assumptions can limit robustness in practice, as real-world acoustic measurements are affected by calibration errors, environmental variability, and noise. These challenges are further exacerbated in SSS, where each one-dimensional ping is associated with a single sensor pose and provides only limited geometric constraints. As a result, reconstruction relies heavily on consistency across multiple pings. This motivates the use of an implicit representation, where bathymetry can be learned as a continuous function from sparse acoustic observations. We adopt a differentiable neural rendering formulation for SSS image formation, where the neural renderer implicitly learns the acoustic patterns directly from data, reducing reliance on analytical models.

Prior work on learning-based bathymetry estimation from SSS trains a convolutional neural network to predict surface normals from SSS images using MBES-derived normals as supervision, and uses the predicted normals to constrain an implicit bathymetric representation~\cite{Xie2023JOE}.
More recently, Xie et al. \cite{Xie2022Sensors,Xie2025JOE} present a neural implicit approach for bathymetry estimation from SSS measurements using gradient-based optimization to disambiguate elevation angles and a Lambertian rendering model to jointly estimate geometry, albedo, and beam profile. These methods either rely on supervised pretraining or fail to account for multiple structures along the elevation arc, leaving ambiguities unresolved and making the optimization prone to local minima.
To address these limitations, our proposed method handles complex structures and resolves the elevation ambiguity.

Also relevant to our work, prior work has studied neural representations for 3D reconstruction using imaging sonar. 
Recent work used neural implicit representation and a differentiable neural renderer to reconstruct 3D underwater objects and to optimize sensor poses~\cite{Qadri2023ICRA, Lin2025IROS}.
Subsequent extensions, NeuSIS-NGP, incorporate neural volume rendering and multi-level hash encoding~\cite{Xie2024RAL}. Related neural approaches have also explored volumetric reconstruction for synthetic aperture sonar \cite{Reed2023TOG} and differentiable space carving for imaging sonar \cite{Feng2024RAL}. 
We instead develop a neural implicit framework for bathymetry estimation from SSS.

%%%%%%%%%%%%%%%%%%%%%%%%%%%%%%%%%%%%%%%%%%%%%%%%%%%%%%%%%%%%%%%%%%%%%%%%%%%%%%%%
\section{Background on Sidescan Sonar Geometry} \label{sec:sss_geometry}

\begin{figure} [th]
    \centering
    \includegraphics[width=\linewidth]{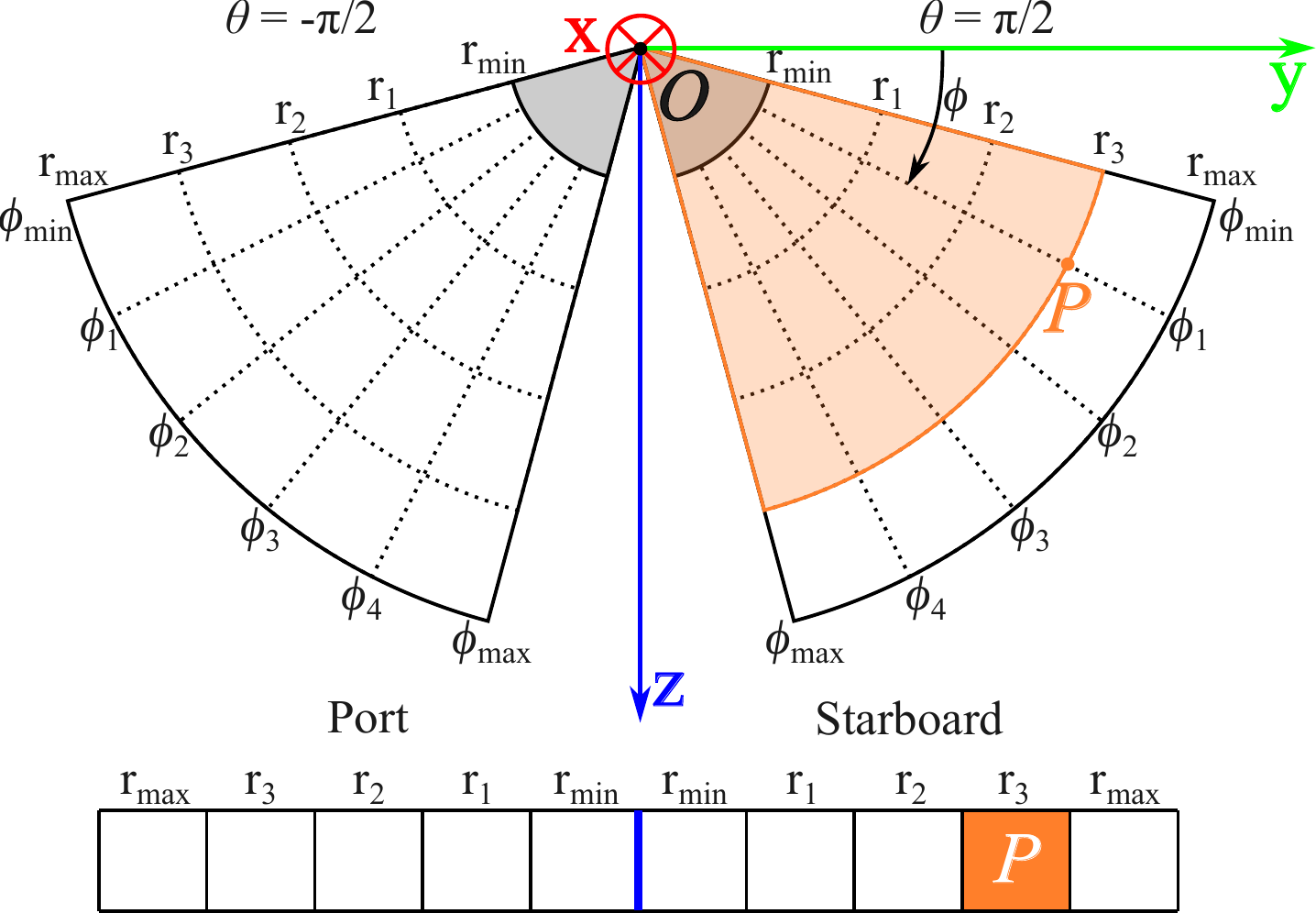}
    \caption{Geometry of a single SSS ping. Under the SSS right-handed coordinate convention, the $x$-axis aligns with the direction of sensor motion and the $y$-axis points toward the starboard side. A 3D point $P$ within the SSS field of view is represented by slant range $r$, azimuth angle $\theta$, and elevation angle $\phi$. The SSS emits acoustic signals to both the port (left) and starboard (right) sides, so the azimuth angle of an object is $-\frac{\pi}{2}$ or $\frac{\pi}{2}$. $r_\text{min}$, $r_\text{max}$, $\phi_\text{min}$, and $\phi_\text{max}$ are the minimum and maximum slant ranges and elevation angles of the SSS. For illustration, only five bins are shown on each side. Point $P$ is projected onto the bin $r_3$ on the starboard side.}
    \label{fig:sss_geometry}
    \vspace{-2em}
\end{figure}

\noindent The SSS emits acoustic signals to the starboard and port sides and measures slant ranges $r$ of reflected acoustic signals based on the time-of-flight~\cite{Sethuraman2025IJRR}. 
Fig. \ref{fig:sss_geometry} illustrates the geometry of the SSS and its associated sensor coordinates.
The $x$-axis, which points into the page, denotes the direction of sensor motion, while the $z$-axis points downward and the $y$-axis points to the starboard side. Each SSS ping produces a one-dimensional intensity profile along the slant range axis for both port and starboard sides. 

Following the standard SSS imaging model~\cite{Sethuraman2025IJRR}, we assume a range-bearing geometry for a 3D point $P(x,y,z)$ in the local sonar Cartesian coordinates and neglect the horizontal beam-width effect, which is typically below $1^\circ$ for conventional SSS systems. The conversions between the Cartesian and the corresponding polar representation $P(r, \theta, \phi)$ are:
\begin{equation}
\begin{aligned}
    \begin{bmatrix}
        x\\ y\\ z
    \end{bmatrix}
    &=
    \begin{bmatrix}
        0\\
        r \sin \theta \cos \phi\\
        r \sin \phi
    \end{bmatrix},
    \;
    \begin{bmatrix}
        \theta\\ r\\ \phi
    \end{bmatrix}
    &\hspace{-0.5em}=
    \begin{bmatrix}
        \arctan2(y,0)\\
        \sqrt{y^2+z^2}\\
        \arctan2(z,|y|)
    \end{bmatrix},
\end{aligned}
\label{eq:sss_coordinate_transform}
\end{equation}

\noindent where $r$, $\theta$, and $\phi$ denote the slant range, azimuth angle, and elevation angle, respectively. In this geometry, because we neglect the horizontal beam width and consider $\theta=\pm \frac{\pi}{2}$, we naturally obtain $x \equiv 0$.
% $x = r \cos\theta \cos\phi = 0$. 
Therefore, $\theta$ is determined by the sign of $y$. We represent a 3D point $P(\mathbf{x})$ by its coordinate vector $\mathbf{x}=(x,y,z)^\top$. Similar to Qadri et al.~\cite{Qadri2023ICRA}, we define $\mathbf{D}(\mathbf{x})$ as the viewing direction from the SSS to $P$ in the world coordinates.

To align with the polar coordinate convention, we reshape each one-dimensional SSS ping by splitting it at the center and arranging it into a two-column matrix. The left and right columns correspond to the port and starboard returns, while the row indices represent slant ranges spanning from the minimum to the maximum measurable range. 

Similar to imaging sonars, SSS cannot resolve the elevation angle of returned signals. As a result, reflections sharing the same azimuth angle and range but different elevation angles are projected onto the same image pixel.

%%%%%%%%%%%%%%%%%%%%%%%%%%%%%%%%%%%%%%%%%%%%%%%%%%%%%%%%%%%%%%%%%%%%%%%%%%%%%%%%
\begin{figure*} [th]
    \centering
    \includegraphics[width=0.83\paperwidth]{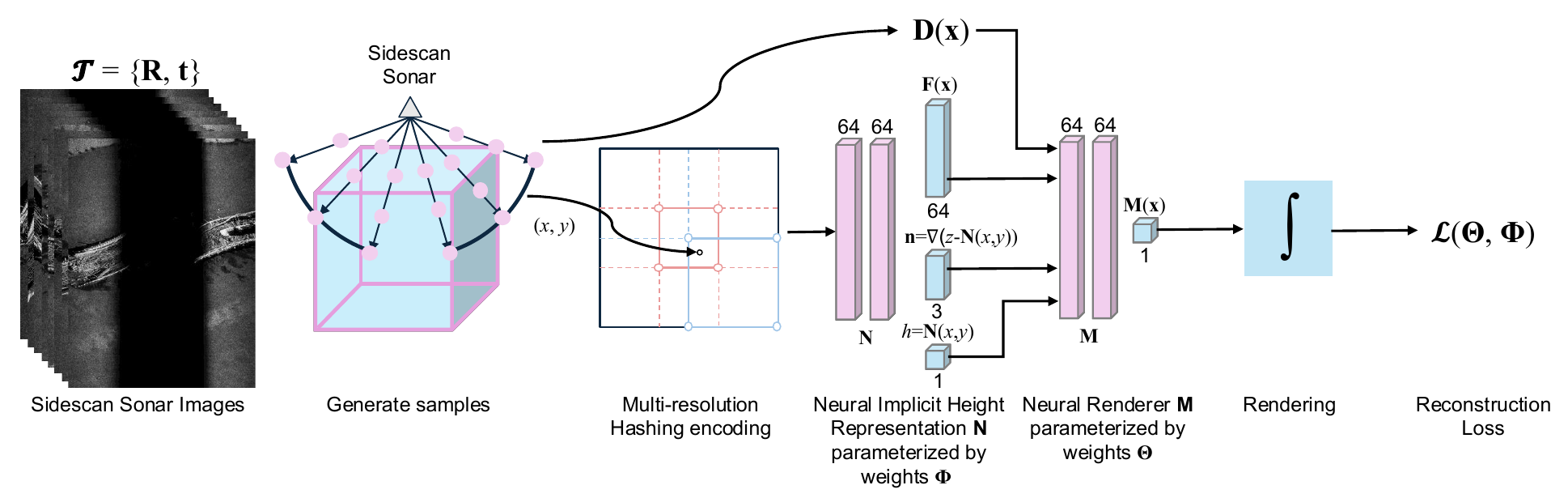}
    \caption{Pipeline overview. Our proposed method optimizes the sonar neural implicit bathymetry networks and renderer by the total rendering loss. The network takes an input set of SSS images with known poses ($\mathcal{T} = \{\textbf{R},\textbf{t}\}$). The sampled 3D points $\mathbf{x}$ and viewing directions $\mathbf{D}(\mathbf{x})$ are encoded by the multi-resolution hash encoder and spherical harmonics encoder. The neural renderer then estimates the acoustic radiance, which is used for the final SSS image formation model. The networks are supervised by the reference SSS images and other constraints included in the reconstruction loss.}
    \label{fig:pipeline}
    \vspace{-1.5em}
\end{figure*}

\section{Technical Approach}

\noindent
Fig. \ref{fig:pipeline} gives an overview of the proposed pipeline, where the SSS image formation model and neural implicit bathymetry network are jointly optimized through the reconstruction loss. The network takes as input a set of SSS images and corresponding poses. For training, we assume access to a sparse prior map measured by an altimeter sensor across the full survey area.

\vspace{-0.5em}
\subsection{Sidescan Sonar Image Formation Model}

\noindent As discussed in Sec. \ref{sec:sss_geometry}, SSS emits acoustic signals and determines range based on the time-of-flight of the reflected signals. The intensity of each pixel is proportional to the cumulative sum of backscattered energy from all reflectors lying along the elevation arc at the same range and azimuth direction. We adopt the general sonar intensity rendering formulation to compute the sonar intensity $I_p=I(r_i,\theta_j)$ at pixel $(r_i, \theta_j)$~\cite{Qadri2023ICRA, Xie2024RAL} :

\vspace{-1em}
\begin{align}
& I_p =  \int_{\phi_\text{min}}^{\phi_\text{max}} \int_{r_i-\epsilon}^{r_i+\epsilon} \beta(\phi) \frac{\textit{E}_e}{r} T(r, \theta_j, \phi) \sigma(r, \theta_j, \phi) \text{d}r\text{d}\phi.
\label{eq:rendering}
\end{align}

\noindent where $\phi_\text{min}, \phi_\text{max}$ denote the minimum and maximum elevation angles, $2\epsilon$ is the patch thickness, and  $E_e$ is the emitted sonar acoustic energy. 
The beam pattern corresponding to the given elevation angle is defined as 
\begin{align}
    \beta(\phi) = \sum_{\mathcal{K}}\Phi_{\mathcal{K}}\exp{(-\frac{(\phi_{\mathcal{K}}-\phi)^2}{2\sigma_\Phi^2})},
    \label{eq:beam_pattern}
\end{align}
where $\Phi_\mathcal{K}$ represents the elevation kernel density weights, which are jointly optimized. $\phi_{\mathcal{K}}$ is the fixed kernel position, and $\sigma_\Phi$ is the kernel spread, set as the range divided by the number of kernels. The transmittance term is defined as
\begin{align}
    T = \exp({-\int_0^{r} \sigma(r', \theta_j, \phi) \text{d}r'}), 
    \label{eq:transmittance}
\end{align}
and $\sigma$ denotes the particle density.

\vspace{-0.5em}
\subsection{Neural Representation}

\noindent  Similar to NeuSIS-NGP~\cite{Xie2024RAL}, the bathymetry is represented as a vertical signed distance function (vSDF), $\Delta (\mathbf{x}) = z-\mathbf{N}(x,y)$, which outputs the difference between the estimated and reference heights given a 3D point $\mathbf{x}(x,y,z)$ in the world coordinates. The neural implicit height representation is defined as $\mathbf{N}$. The surface normal $\mathbf{n}$ is calculated from the gradients of the height representation $\mathbf{N}$ with respect to its input positions:
\begin{align}
    \mathbf{n} = [-\nabla_x \mathbf{N}(x, y), -\nabla_y \mathbf{N}(x, y), 1].
\end{align}

\noindent The multi-resolution hash encoder encodes the positions and outputs the feature vectors $\mathbf{F}$. Neural renderer $\mathbf{M}$ computes the outgoing acoustic radiance at $\mathbf{x}$  to approximate the SSS image formation integral $\hat{I}_p$ at each pixel. 

\begin{figure} [th]
    \centering
    \includegraphics[width=\linewidth]{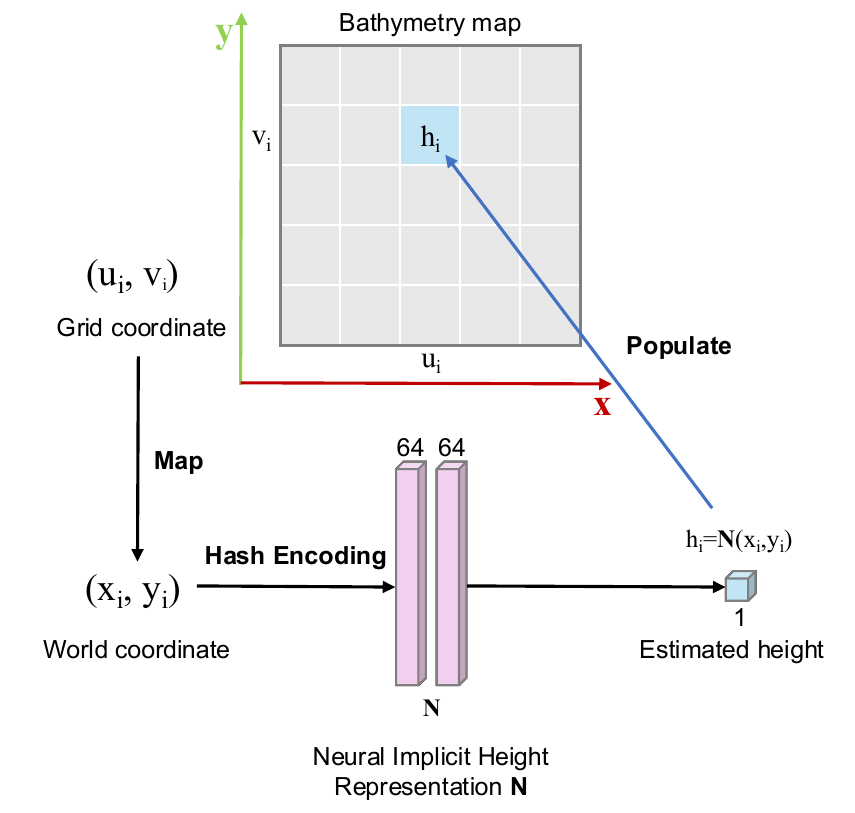}
    \caption{Process of bathymetry map generation. The bathymetry matrix is constructed by populating each grid with the height estimated from the neural implicit height representation evaluated at its corresponding geographic location. Grid coordinates of the bathymetry map are in one-to-one correspondence with geographic positions.}
    \label{fig:bathymetry_generation}
    \vspace{-3em}
\end{figure}

The neural implicit height network, $\mathbf{N}, $ supports bathymetry map generation at arbitrary resolutions. Fig. \ref{fig:bathymetry_generation} shows the bathymetry map generation process. The grid resolution, together with the geographic bounds of the map (i.e., minimum and maximum $x$- and  $y$-coordinates, measured in meters), defines a one-to-one mapping between coordinates $(u_i, v_i)$ of the map grid $i$ and world coordinates $(x_i, y_i)$. The bathymetry map is then generated by evaluating the neural implicit height representation at each location and populating the predicted height values to their corresponding grid positions.

\vspace{-0.3em}
\subsection{Discretization of the Image Formation Integral}

\noindent We discretize and approximate Eq.~\ref{eq:rendering} by sampling 3D points along both acoustic rays and arcs:
\vspace{-0.2em}
\begin{align}
    &    \hat{I}_p = \sum_{\mathbf{x} \in \mathcal{A}_p}
    \frac{1}{r(\mathbf{x})} \beta[\mathbf{x}] T[\mathbf{x}] \alpha[\mathbf{x}] \mathbf{M}(\mathbf{x})
    \label{discreteImageFormationModelSonar}
\end{align}
where $\mathcal{A}_{p}$ is the set of sampled 3D points in the world coordinates along the acoustic elevation arc at pixel $p(r_i, \theta_i)$ and $ \mathbf{M}$ is the estimated radiance at $\mathbf{x}$. The discrete transmittance $T[\mathbf{x}]$ and opacity $\alpha[\mathbf{x}]$ terms are computed by sampling along acoustic rays shooting out from the sensor to the sampled 3D points in $\mathcal{A}_{p}$.
For each spatial sample \(\mathbf{x}\) along the acoustic ray, the discretized opacity at \(\mathbf{x}\) is defined using the next sample $\bar{\mathbf{x}}$:
 \begin{align}
    \alpha[\mathbf{x}] = \max\left(\frac{\mathbf{\Phi}_s(\Delta(\mathbf{x}))-\mathbf{\Phi}_s(\Delta(\bar{\mathbf{x}}))}{\mathbf{\Phi}_s(\Delta(\mathbf{x}))}, 0\right),
    \label{discreteopacity}
\end{align} 
where $\mathbf{\Phi}_s(x) = (1+e^{-sx})^{-1}$ is the sigmoid function and $s$ is a trainable standard deviation parameter while the discrete transmittance is modeled as 
\begin{align}
    T[\mathbf{x}] = \prod_{\mathbf{x_r} \in \mathcal{R}_{p} \backslash \{\mathbf{x}\} } (1-  \alpha[\mathbf{x_r}]),
\end{align}
where $\mathcal{R}_{p}$ denotes the set of points sampled along the acoustic ray that ends at $x$.
We also adopt the hierarchical sampling technique based on the trainable standard deviation $s$, which was introduced by NeuSIS-NGP~\cite{Xie2024RAL}.
\vspace{-1.5em}

%%%%%%%%%%%%%%%%%%%%%%%%%%%%%%%%%%%%%%%%%%%%%%%%%%%%%%%%%%%%%%%%%%%%%%%%%%%%%%%%

\subsection{Loss Function}
\vspace{-0.5em}

\noindent We define the following set of trainable parameters: $\Phi$ are the weights of the neural height network $\mathbf{N}$, and $\Theta$ are the weights of the neural renderer $\mathbf{M}$. Our loss function is composed of three terms.
The intensity loss 
\begin{align}
\mathcal{L}_{\text{int}} = \frac{1}{|\mathcal{P}|} \sum_{p \in \mathcal{P}}\Vert\hat{I}_p(\Theta, \Phi) - I_p\Vert_1,
\end{align}
where $\mathcal{P}$ denotes the set of sampled pixels, encourages the predicted intensity to be consistent with the observed intensity at pixel $p$ in the training SSS images. The eikonal loss \cite{gropp2020implicit}
\begin{align}
    \mathcal{L}_{\text{eik}} = \frac{1}{|\mathcal{A}_{p}\bigcup\mathcal{R}_{p}|} \sum_{\mathbf{x} \in \mathcal{A}_{p}\bigcup\mathcal{R}_{p} } (\Vert \mathbf{n}(\mathbf{x})\Vert_2 - 1)^2,    
\end{align}
encourages the vSDF to produce smoother bathymetry maps. And finally the altimetry loss facilitates the convergence of models:
\begin{align}
    \mathcal{L}_{\text{alt}} =\frac{1}{|\mathbf{X}_\text{alt}|} \sum_{\spatialpoint \in \mathbf{X}_\text{alt}} \Vert \Delta (\mathbf{x}) \Vert_1,
\end{align}
where $\mathbf{X}_\text{alt}$ denotes a sparse prior map measured by the altimetry sensor  across the full area.
The final training loss term is:
\begin{align}
\mathcal{L}(\Theta, \Phi) = \lambda_{\text{int}} \mathcal{L}_{\text{int}} + \lambda_{\text{eik}} \mathcal{L}_{\text{eik}} + \lambda_{\text{alt}} \mathcal{L}_{\text{alt}}.
\label{lossfunction}
\end{align}
where $\lambda_{\text{int}}, \lambda_{\text{eik}}, \lambda_{\text{alt}}$ are tunable scaling factors for the intensity loss, eikonal loss and altimetry loss, respectively. Our objective function is therefore: 
\begin{align}
    \Theta^*, \Phi^* = \argmin_{\Theta, \Phi} \mathcal{L}(\Theta, \Phi)
    \label{optimproblem}
\end{align}

%%%%%%%%%%%%%%%%%%%%%%%%%%%%%%%%%%%%%%%%%%%%%%%%%%%%%%%%%%%%%%%%%%%%%%%%%%%%%%%%

\begin{figure*} [ht!]
    \centering
    \includegraphics[width=0.80\paperwidth]{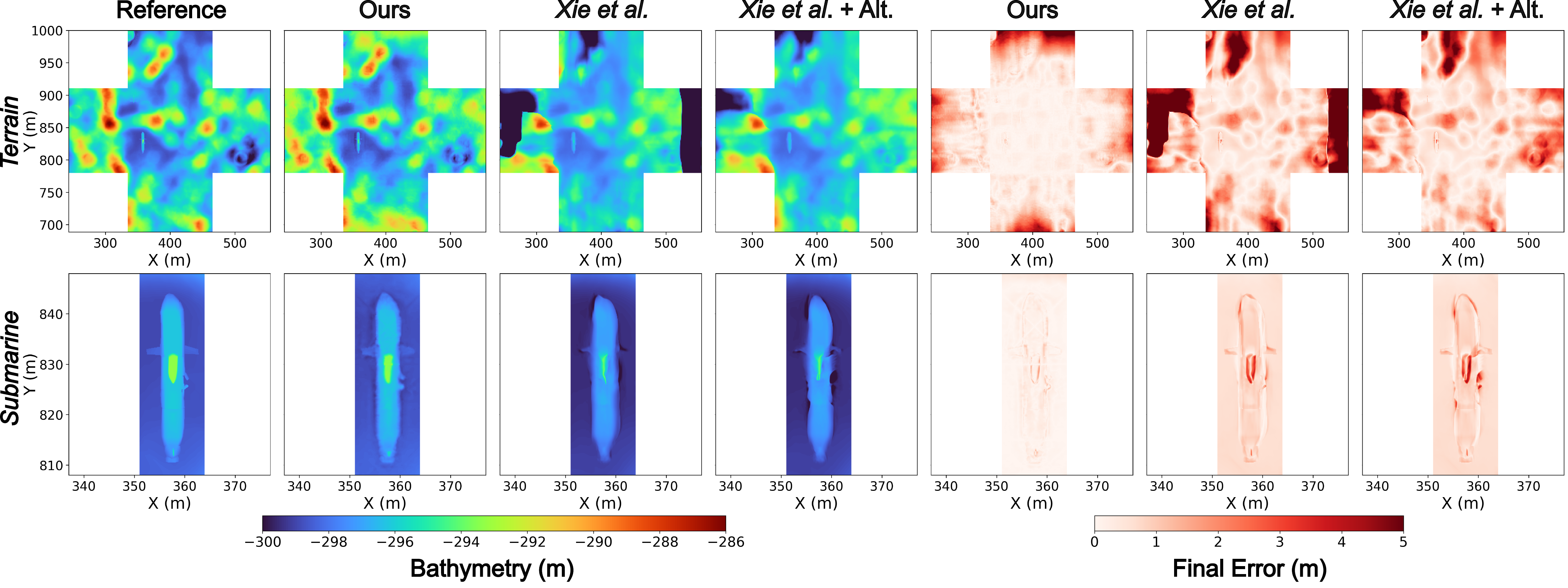}
    \caption{Visualization of bathymetry estimation results for synthetic datasets. The axes $x$ and $y$ are expressed in meters in the world coordinate frame. From all the reconstructed maps and the corresponding error maps, our method produces more consistent estimations and achieves lower reconstruction errors relative to the reference maps compared to Xie et al.~\cite{Xie2022Sensors} and the modified version with sparse altimetry priors.}
    \label{fig:synthetic_bathymetry}
    \vspace{-1.5em}
\end{figure*}

\section{Experiments and Results}

\noindent  We trained our models on a 96GB NVIDIA RTX PRO 6000 Blackwell Max-Q GPU with an AMD Ryzen Threadripper PRO 7975WX CPU. Each experiment ran for approximately 100k to 300k iterations, depending on the number of sonar frames in the dataset, with an initial learning rate of 0.01 that linearly decayed by factors of 0.97 and 0.9 for the simulated and real-world datasets, respectively. Table \ref{table:num_frames} summarizes the number of surveys, poses, and iterations in each real and simulated (`Sim.') dataset.
For quantitative evaluation of bathymetry estimation, we report mean absolute error (MAE) and standard deviation (STD) between the estimated and ground-truth bathymetry maps.

\begin{table}[h!]
    \vspace{-0.5em}
    \begin{center}
    \resizebox{\linewidth}{!}{
        \begin{tabular}{c|c|c|c|c}
            \toprule
            Datasets & Sim. Terrain & Sim. Submarine & \textit{D.M. Wilson} & \textit{Montana} \\
            \hline
            Number of Survey Legs     & 21     & 8      & 8     & 8     \\ 
            Number of Training Legs     & 20     & 7      & 7     & 7     \\ 
            Number of Poses       & 10000  & 4900   & 7000  & 10500 \\ 
            Number of Iterations  & 300000 & 147000 & 210000 & 315000 \\ 
            \cline{1-3}
            \hline
    
        \end{tabular}
    }
    \end{center}
    \vspace{-0.5em}
    \caption{Dataset statistics and training configurations, including survey legs, training legs, poses, and iterations. \textit{Terrain} and \textit{Submarine} datasets are collected from HoloOcean simulator, while \textit{D.M. Wilson} and \textit{Montana} are real-world datasets.}
    \label{table:num_frames}
    \vspace{-1.5em}
\end{table}

%%%%%%%%%%%%%%%%%%%%%%%%%%%%%%%%%%%%%%%%%%%%%%%%%%%%%%%%%%%%%%%%%%%%%%%%%%%%%%%%

\subsection{Simulation}

\noindent We generate two synthetic SSS datasets using HoloOcean \cite{Potokar24JOE}, a realistic underwater simulator. The generated SSS images are free of noise. The datasets represent two scenarios: a large-scale terrain survey and a submarine resting on the seafloor. The SSS ranges are set to 100 m and 30 m for the terrain and submarine datasets, respectively, with an elevation angle of 130 degrees in both cases. The simulated AUV is equipped with a Doppler Velocity Log (DVL), an inertial measurement unit (IMU), and a bathymetry sensor that measures the altimetry directly beneath the vehicle.

\begin{figure} [hbtp!]
    \centering
    \includegraphics[width=0.7\linewidth]{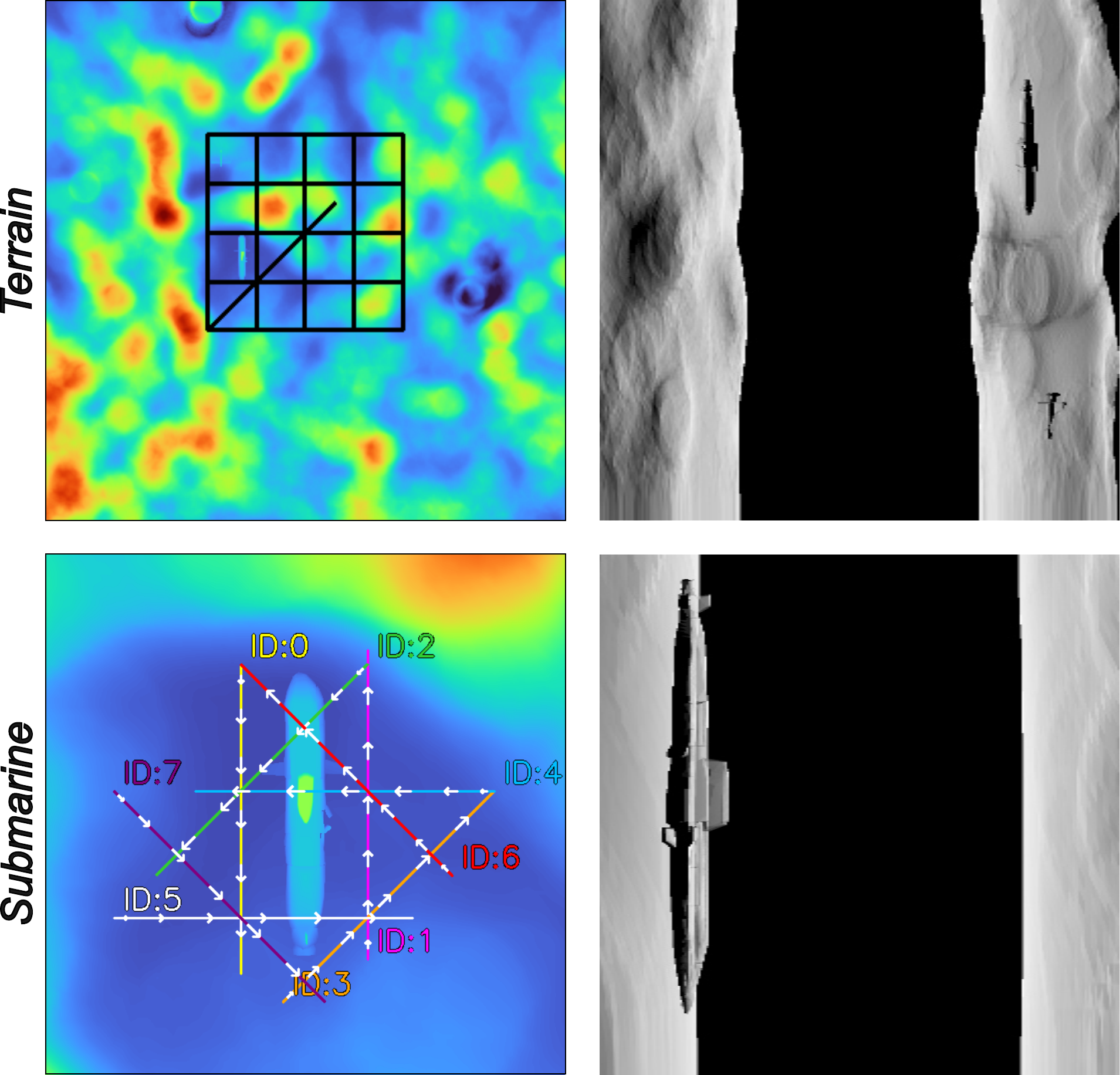}
    \caption{Survey trajectories and example SSS images of the simulated \textit{Terrain} and \textit{Submarine} datasets. The \textit{Terrain} dataset consists of 20 training survey legs and 1 northeast-oriented survey leg for validation. The \textit{Submarine} dataset contains 8 survey legs following an OID pattern.}
    \label{fig:synthetic_trajectory}
    \vspace{-2em}
\end{figure}

For data generation, the terrain dataset follows a grid-pattern trajectory, consisting of five survey legs in each of the north, south, east, and west directions. The submarine dataset uses an object identification (OID) trajectory pattern. Fig. \ref{fig:synthetic_trajectory} illustrates the survey trajectories and example SSS images for both the simulated terrain and simulated submarine datasets.

\begin{figure*} [ht!]
    \centering
    \includegraphics[width=0.83\paperwidth]{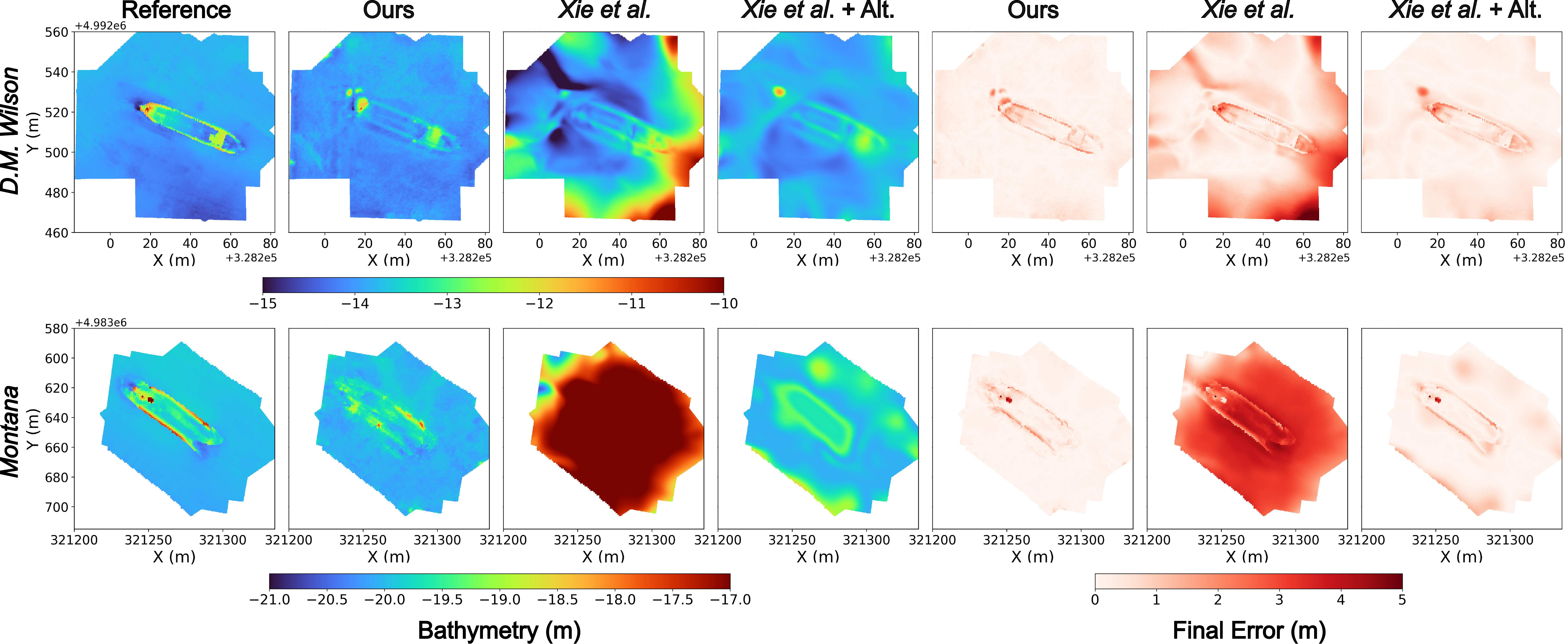}
    \caption{Visualization of bathymetry estimation results for real-world datasets. The axes $x$ and $y$ are expressed in meters in the WGS 84 / UTM Zone 17N coordinate system. Despite the presence of pose errors and noisy SSS measurements, our method produces more accurate reconstructions of the shipwrecks and achieves lower errors compared to both variants of Xie et al.~\cite{Xie2022Sensors}.}
    \label{fig:real_bathymetry}
    \vspace{-1em}
\end{figure*}

Fig. \ref{fig:synthetic_bathymetry} and Table \ref{table:synthetic_bathymetry_metrics} demonstrate the qualitative and quantitative bathymetry estimation and error maps for both simulated datasets. The elevation beam kernels are disabled because HoloOcean uniformly accumulates reflected acoustic energy along the elevation arcs. We compare our proposed method with Xie et al.~\cite{Xie2022Sensors} and its modified version with sparse altimetry (`Alt.') priors. Quantitatively, our proposed method produces accurate and consistent bathymetry estimations, while the baselines cannot reconstruct the actual shape of the underwater terrains or objects.

\begin{table}[b!]
\vspace{-2.5em}
\begin{center}
\resizebox{0.8\linewidth}{!}
{
\begin{tabular}{c|cc|cc}
\toprule
\multirow{2}{*}{\backslashbox{Methods}{Dataset}}
& \multicolumn{2}{c|}{\textit{Terrain}}
& \multicolumn{2}{c}{\textit{Submarine}} \\
\cline{2-5}
& MAE$\downarrow$ & STD$\downarrow$
& MAE$\downarrow$ & STD$\downarrow$ \\
\hline
\textit{Xie et al.}
& 1.911 & 2.522 & 0.674 & 0.328 \\
\textit{Xie et al.} + Alt.
& 1.205 & 1.244 & 0.713 & 0.401 \\
Ours
& \textbf{0.727} & \textbf{0.973} & \textbf{0.117} & \textbf{0.158} \\
\hline
\end{tabular}
}
\end{center}
\vspace{-0.8em}
\caption{Bathymetry estimation metrics (m) for simulated datasets.}
\label{table:synthetic_bathymetry_metrics}
\vspace{-1em}
\end{table}

\subsection{Real-world Experiment}
\noindent We validate our proposed method on real-world datasets previously collected at the Thunder Bay National Marine Sanctuary in Lake Huron, Michigan. The experiments were conducted using the Michigan Technological University (MTU) Iver3 AUV, equipped with an EdgeTech 2205 dual-frequency (540/1610 kHz) SSS, a 3D bathymetric system, and a Teledyne RDI Explorer DVL which provides long-term accuracy of $\pm 0.3\% \pm 0.2 \text{ cm/s}$ \cite{Sethuraman2025IJRR}.

\begin{figure} [ht]
    \centering
    \includegraphics[width=0.7\linewidth]{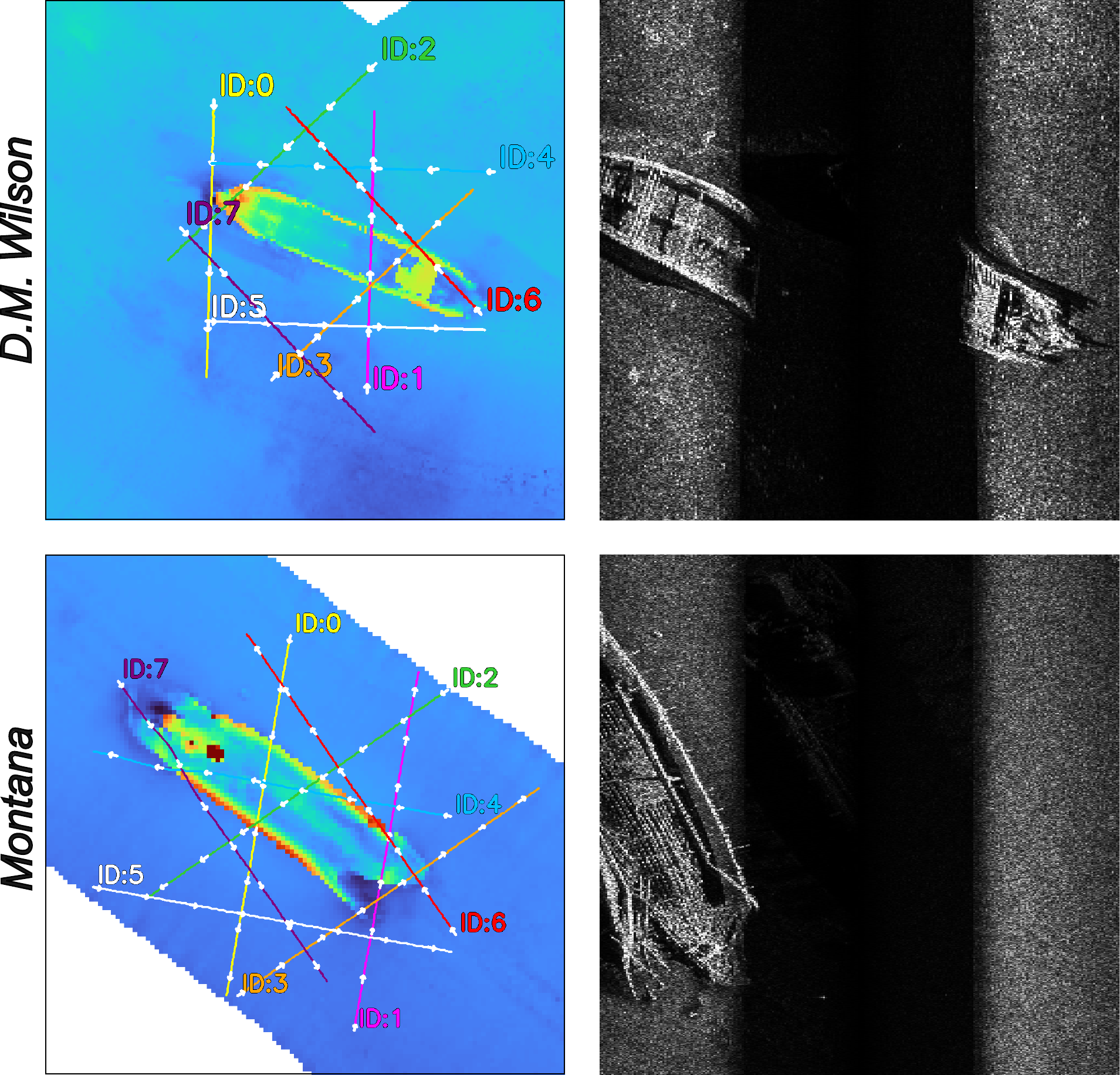}
    \caption{Survey trajectories and example SSS images of the \textit{D.M. Wilson} and \textit{Montana} datasets. Both datasets contain 8 survey legs following an OID pattern.}
    \label{fig:real_trajectory}
    \vspace{-1em}
\end{figure}

We evaluate our method on the \textit{D.M. Wilson} and \textit{Montana} shipwreck datasets.  Fig. \ref{fig:real_trajectory} shows the trajectories and example SSS of both real sites. The sonar range is set to $30$ meters with an elevation aperture of $120^\circ$. For each bathymetry estimate, we apply a mask to regions with non-zero covisibility. For real-world datasets, we enable 20 elevation beam kernels distributed along the elevation arcs. Fig. \ref{fig:real_bathymetry} and Table \ref{table:real_bathymetry_metrics} present the qualitative and quantitative results on real-world data.

Our method is able to reconstruct the overall structure of the shipwreck within the SSS field of view, despite misalignment and scale discrepancies. These errors are likely caused by noisy and drifting AUV pose estimates, as well as non-trivial noise and artifacts present in real SSS measurements. Such factors make the real-world setting more challenging than simulation, leading to degraded reconstruction quality. Nevertheless, compared to both the original and altimetry-augmented versions of Xie et al.~\cite{Xie2022Sensors}, our method still demonstrates promising performance for bathymetry estimation in real-world scenarios. 

\begin{table}[h!]
\vspace{-0.5em}
\begin{center}
\resizebox{0.8\linewidth}{!}
{
\begin{tabular}{c|cc|cc}
\toprule
\multirow{2}{*}{\backslashbox{Methods}{Dataset}}
& \multicolumn{2}{c|}{\textit{D.M. Wilson}}
& \multicolumn{2}{c}{\textit{Montana}} \\
\cline{2-5}
& MAE$\downarrow$ & STD$\downarrow$
& MAE$\downarrow$ & STD$\downarrow$ \\
\hline
\textit{Xie et al.}
& 0.890 & 0.905 & 2.880 & 0.812 \\
\textit{Xie et al.} + Alt.
& 0.266 & 0.295 & 0.253 & 0.328 \\
Ours
& \textbf{0.173} & \textbf{0.242} & \textbf{0.130} & \textbf{0.270} \\
\hline
\end{tabular}
}
\end{center}
\vspace{-0.5em}
\caption{Bathymetry estimation metrics (m) for real-world datasets.}
\label{table:real_bathymetry_metrics}
\vspace{-1.5em}
\end{table}

%%%%%%%%%%%%%%%%%%%%%%%%%%%%%%%%%%%%%%%%%%%%%%%%%%%%%%%%%%%%%%%%%%%%%%%%%%%%%%%%
\section{Conclusion and Future Work}
\noindent 
We proposed a neural implicit reconstruction framework for bathymetry estimation using SSS imagery. Through experiments on both synthetic and real-world datasets, we demonstrated that our method achieves improved accuracy on bathymetry estimation compared to the existing approach.

For future work, we see multiple promising directions. First, we will explore incorporating sonar noise and pose error in
the neural rendering pipeline to improve results on real data. Second, since our current process handles one sonar ping at a time, we will transition from single-ping processing to batch training to fully leverage our GPU hardware. This will enable parallel processing of multiple pings, boosting data throughput to work toward an efficient training pipeline for bathymetry estimation from SSS data.

%%%%%%%%%%%%%%%%%%%%%%%%%%%%%%%%%%%%%%%%%%%%%%%%%%%%%%%%%%%%%%%%%%%%%%%%%%%%%%%%
% \section{Acknowledgment}
% \noindent

%%%%%%%%%%%%%%%%%%%%%%%%%%%%%%%%%%%%%%%%%%%%%%%%%%%%%%%%%%%%%%%%%%%%%%%%%%%%%%%%
% \section*{Appendix}
\bibliographystyle{IEEEtran}
\IEEEtriggeratref{15}
\bibliography{main}

\end{document}